%% file: egpaper_final.tex
\pdfoutput=1
\documentclass[10pt,twocolumn]{article}

\usepackage{cvpr}
\usepackage{times}
\usepackage{epsfig}
\usepackage{graphicx}
\usepackage{amsmath}
\usepackage{amssymb}

\usepackage[caption=false]{subfig}

\cvprfinalcopy 

\def\cvprPaperID{****} 

\begin{document}

\title{Deepfake Detection using Spatiotemporal Convolutional Networks}


 \author{Oscar de Lima, Sean Franklin, Shreshtha Basu, Blake Karwoski, Annet George\\
University of Michigan\\
Ann Arbor, MI 48109\\
oidelima, sfrankl, shreyb, bkarw, annetbge@umich.edu}


\maketitle

\begin{abstract}
   Better generative models and larger datasets have led to more realistic fake videos that can fool the human eye but produce temporal and spatial artifacts that deep learning approaches can detect. Most current Deepfake detection methods only use individual video frames and therefore fail to learn from temporal information. We created a benchmark of the performance of spatiotemporal convolutional methods using the Celeb-DF dataset. Our methods outperformed state-of-the-art frame-based detection methods. Code for our paper is publicly available at {https://github.com/oidelima/Deepfake-Detection}.
\end{abstract}

\section{Introduction}

In recent years, Deepfakes (manipulated videos) have become an increasing threat to social security, privacy and democracy. As a result, research into the detection of such Deepfake videos has taken many different approaches. Initial methods tried to exploit discrepancies in the fake video generation process, however more recently, research has moved toward using deep learning approaches for this task. 
\par
From a larger perspective, Deepfake detection can be considered a binary classification problem to distinguish between `real' and `fake' videos. There are many architectures that have achieved remarkable results and these are mentioned in the following section. However, most of these methods rely only on information present in a single image, performing analysis frame by frame, and fail to leverage temporal information in the videos. An area of research that has delved deeper into using information across frames in a video is `Action-Recognition'. 
\par
In this paper, we aim to apply techniques used for video classification, that take advantage of 3D input, on the Deepfake classification problem at hand. All the convolutional networks we tested (apart from RCN) were pre-trained on the Kinetics dataset \cite{kay2017kinetics},  a large scale video classification dataset. The methods were analysed using video clips of fixed lengths.
\par
For this work we have chosen the Celeb-DF (v2) dataset \cite{li2019celeb} which contains 590 real videos collected from YouTube and 5639 corresponding synthesised videos of high quality. Celeb-DF was selected as it has proved to be a challenging dataset for existing Deepfake detection methods due to its fewer noticeable visual artifacts in synthesised videos. 

\section{Related Work}

In the last few decades, many methods for generating realistic synthetic faces have surfaced \cite{dale2011video, garrido2014automatic,  thies2018facevr,thies2016face2face,  korshunova2017fast, bregler1997video,thies2015real,suwajanakorn2017synthesizing,averbuch2017bringing,suwajanakorn2015makes,pham2018generative}. Most of the early methods fell out of use and were replaced by generative adversarial networks and style transfer techniques \cite{FakeApp246:online, dfakerdf55:online,iperovDe80:online,shaoanlu7:online}. Li et al. \cite{li2019celeb} proposed the Deepfake maker method which detected faces from an input video, extracted facial landmarks, aligned the faces to a standard configuration, cropped the faces and fed them to an encoder-decoder to generate faces of a person with a target's expression on it.

The data used in these algorithms have come from several large-scale DeepFake video datasets such as FaceForensics++ \cite{rossler2019faceforensics++}, DFDC \cite{dolhansky2019deepfake}, DF-TIMIT \cite{korshunov2018deepfakes}, UADFV \cite{yang2019exposing}. The Celeb-DF Dataset \cite{li2019celeb} claimed that it was more realistic than the other datasets because the sythesis method they used led to less visual artifacts. Celef-DF has 590 real videos and 5,639 fake ones. Their average length is 13 seconds with a frame rate of 30 fps. The videos were gathered from YouTube corresponding to interviews from 59 celebrities. The synthetic videos in Celeb-DF were made with the DeepFake maker algorithm \cite{li2019celeb}. The resolution of the videos in Celeb-DF was 256 x 256 pixels.

The importance of the task of detecting Deepfakes has led to the development of numerous methods. Rossler et al. \cite{rossler2019faceforensics++} proposed the XceptionNet model that was trained on the Faceforensics++ dataset. Other popular Deepfake detection approaches include Two-Stream \cite{zhou2017two}, MesoNet \cite{afchar2018mesonet}, Headpose \cite{yang2019exposing}, FWA \cite{li2018exposing}, VA \cite{matern2019exploiting}, Multi-task \cite{nguyen2019multi}, capsule \cite{nguyen2019use} and DSP-FWA \cite{he2015spatial}.

Li et al. \cite{li2019celeb} tested all the these methods on the Celeb-DF but didn't train on them. Kumar and Bhavsar \cite{kumar2020detecting}, on the other hand, trained and tested on Celeb-df using the XceptionNet architecture with metric learning. 

Even though video classification methods using spatio-temporal features haven't seen the stratospheric success of deep learning based image classification, several methods have seen some success such as C3D \cite{tran2018closer}, Recurrent CNN \cite{guera2018deepfake}, ResNet-3D \cite{hara2017learning}, ResNet Mixed 3D-2D \cite{tran2018closer}, Resnet (2+1)D \cite{tran2018closer} and I3D \cite{carreira2017quo}.


\section{Method}
\input{Method.tex}

\section{Experiments}
We evaluate all our methods by measuring the top test accuracy and the top ROC-AUC scores. To avoid errors related to numerical imprecision the scores are rounded to 4 decimal points. 

\subsection{Baselines}

We are comparing the performance of our video-based methods against a selection of methods that only work on the level of frames and don't learn from temporal information Table \ref{table:frame}. Li et al. \cite{li2019celeb} made a benchmark on the ROC-AUC scores of frame based methods tested on Celeb-DF but trained on other Deepfake datesets.
\par
More recently, Kumar and Bhavsar \cite{kumar2020detecting} used metric learning using Xception in a triplet network architecture to make the detections. This method was trained and tested on Celeb-DF making it a fairer comparison for our spatio-temporal methods. Additionally, Durall et al. \cite{durall2019unmasking} proposed a method based on running a linear classifier on simple features from the discrete fourier transform of the frames of the videos. We re-implemented that method, and trained it on Celeb-DF.

\subsection{Result and analysis}

We tested some of the most popular networks that take advantage of temporal features. All the networks were trained on the Celeb-DF dataset starting from the pretrained published weights. No layers were frozen for training. Each method was trained for over 25 epochs and the best ROC-AUC score and test accuracies were recorded in Table \ref{table:Video}.
\par
The learning rate was set to start at 0.001 and be divided by 10 every 10 epochs. The optimizer used was stochastic gradient descent with a momentum of 0.9 and a weight decay of 0.0005. The criterion used was cross entropy loss. To account for the imbalance between positive and negative samples in the training set, the criterion weighted each class inversely proportional to the number of samples they had in the training set.

\begin{figure}
\begin{tabular}{cc}
\includegraphics[width=1\columnwidth]{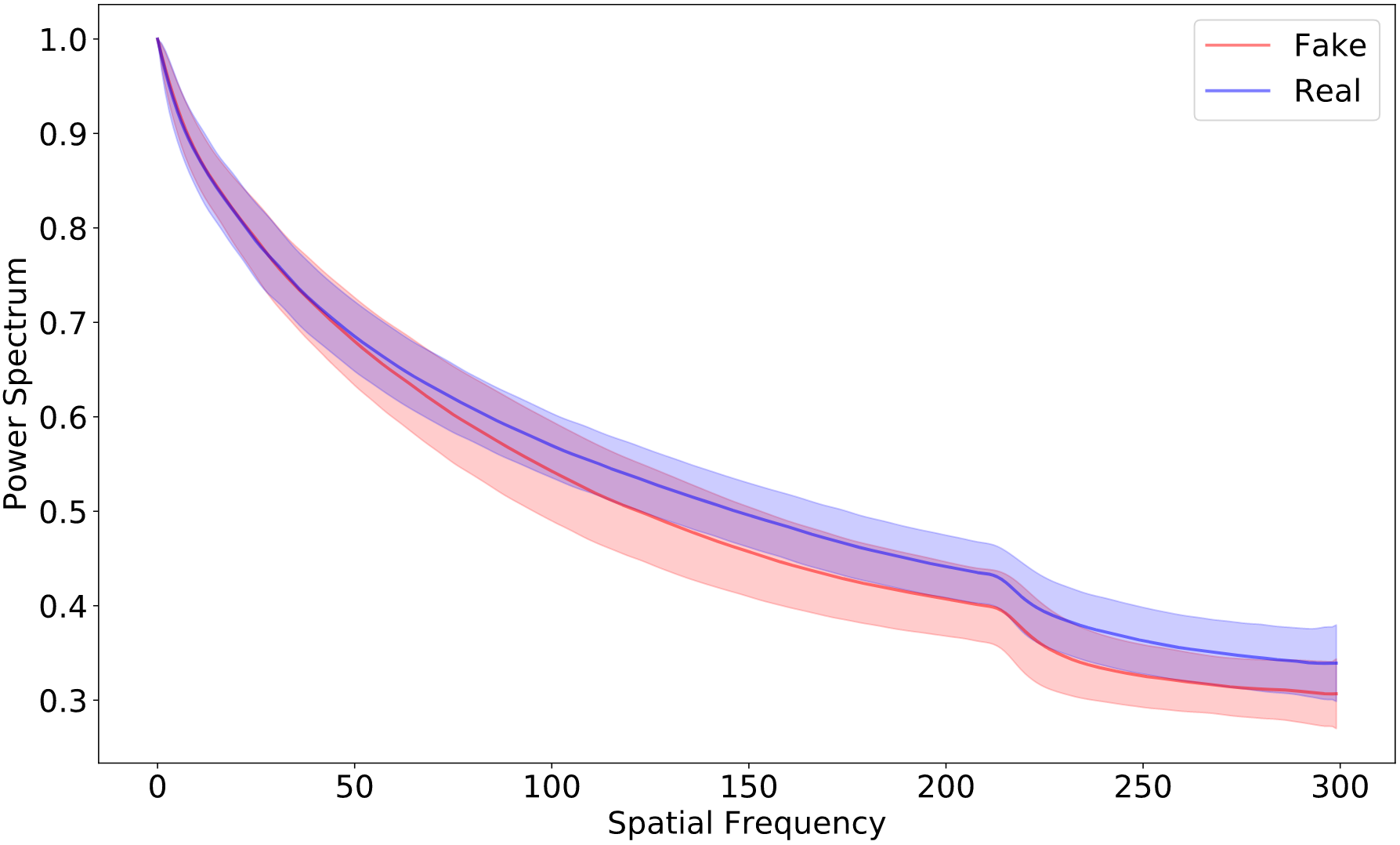}\\
\includegraphics[width=1\columnwidth]{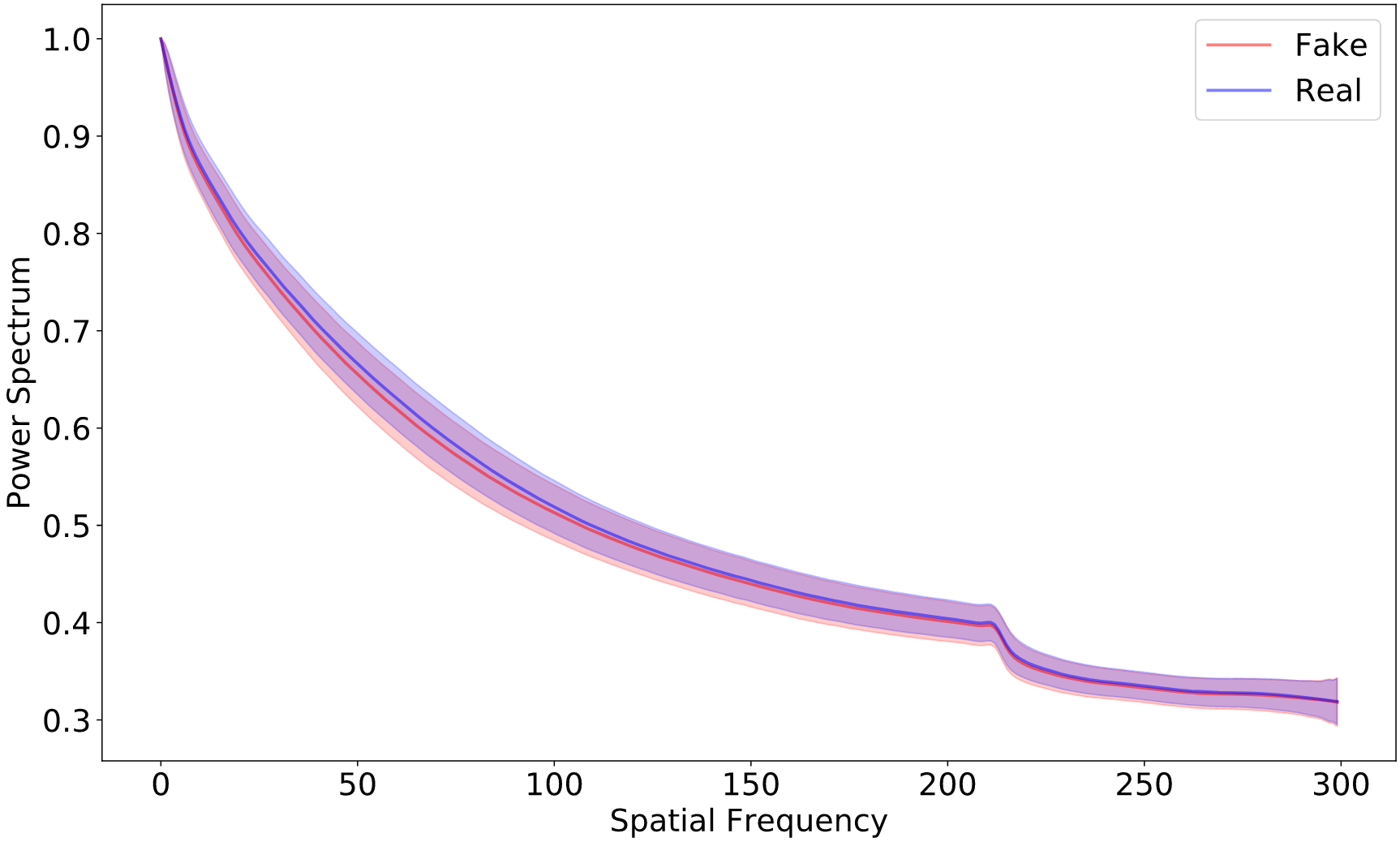}\\
\end{tabular}
\caption{Comparison of power spectra means and standard deviations for real and fake images in the FaceForensics (above) and Celeb-DF datasets.}
\label{DFT_frequency}
\end{figure}

The classical frame based method tested was able to achieve an accuracy of only 66.8\%. This is likely due to the reduced statistical discrepancy between the real and fake images in Celeb-DF versus FaceForensics++. As shown in the plots in figure \ref{DFT_frequency}, for the cropped Celeb-DF the relative power for each frequency remains within one standard deviation of the mean between fake and real, unlike the FaceForensics++ results. This lack of differentiation between the real and fake statistics can explain the lower performance of the classifier on this dataset.
\vspace{0.5cm}

\begin{table}
\begin{center}
\begin{tabular}{|l|c|}
\hline
Method & ROC-AUC \% \\
\hline\hline
Two-Stream* \cite{zhou2017two} & 53.8 \\
Meso4* \cite{afchar2018mesonet}& 54.8 \\
MesoInception4* \cite{afchar2018mesonet}& 53.6\\
HeadPose* \cite{yang2019exposing}& 54.6\\
FWA* \cite{li2018exposing} & 56.9 \\
VA-MLP* \cite{matern2019exploiting} & 55.0 \\
VA-LogReg* \cite{matern2019exploiting}& 55.1\\
Xception-raw* \cite{rossler2019faceforensics++}& 48.2\\
Xception-c23* \cite{rossler2019faceforensics++}& 65.3\\
Xception-c40* \cite{rossler2019faceforensics++}& 65.5\\
Multi-task* \cite{nguyen2019multi}& 54.3\\
Capsule* \cite{nguyen2019use}& 57.5\\
DSP-FWA* \cite{he2015spatial}& 64.6\\
DFT \cite{durall2019unmasking}& 66.8\\
Xception-metric-learning \cite{kumar2020detecting}& 99.2\\
\hline
\end{tabular}
\end{center}
\caption{ROC-AUC Scores for different baseline frame-level deepfake detection methods on Celeb-DF. Methods with * were not trained on Celeb-DF}
\label{table:frame}
\end{table}

\begin{table}
\begin{center}
\begin{tabular}{|l|c|c|}
\hline
Method & ROC-AUC \% & Accuracy \%\\
\hline\hline
RCN  \cite{guera2018deepfake}& 74.87 & 76.25\\
R2Plus1D \cite{tran2018closer}& 99.43 & 98.07\\
I3D \cite{carreira2017quo}& 97.59 & 92.28\\
MC3 \cite{tran2018closer}& 99.30 & 97.49\\
R3D \cite{hara2017learning}& 99.73 & 98.26\\
\hline
\end{tabular}
\end{center}
\caption{Best test ROC-AUC Scores and Accuracies for the spatio-temporal convolutional methods trained on Celeb-DF.}
\label{table:Video}
\end{table}

\begin{figure}
 \includegraphics[width=1\columnwidth,height=6cm]
 {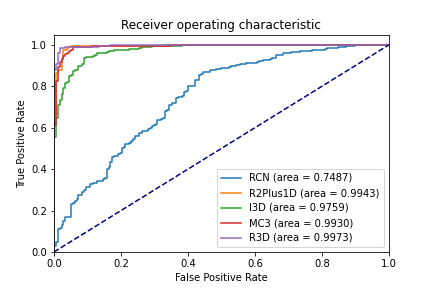} 
\caption{ROC Curves for Spatio-temporal convolutional methods.}
\end{figure}

\begin{figure}
 \includegraphics[width=1\columnwidth,height=6cm]
 {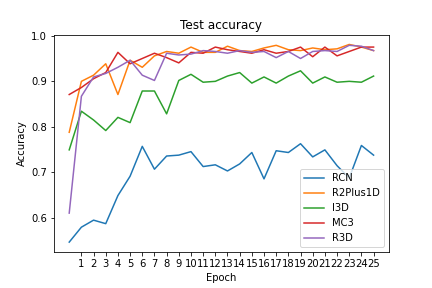} 
\caption{Top test accuracies for Spatio-temporal convolutional methods.}
\end{figure}

\section{Conclusions}
In this paper, we describe and evaluate the efficacy of action recognition methods to detect AI-generated Deepfakes. Our methods differ from previously explored methods because the networks make decisions while incorporating temporal information. This extra information helped several of these networks beat the state of the art baseline frame-based methods. In particular, R3D outperformed the other networks, even I3D \cite{carreira2017quo} which was better at action recognition. We hope that this paper will help future researchers in discovering effective ways of detecting Deepfakes.

{\small
\bibliographystyle{ieee_fullname}
\bibliography{egbib}
}

\end{document}

%% file: Method.tex
In this section we outline the different network architectures we used to perform DeepFake classification. Random cropping and temporal jittering is performed on all methods.

\subsection{Pre-Processing}
The dataset we used was Celeb-DF V2. It contains 590 real videos, and 5639 fake videos. To pre-process all the videos we decided it would best to remove information that might distract our net from learning what was important. 
\par
In a synthesized video, the only part of the frame that is synthesized is over top of the face, so like the frame-based deep fake detection methods, we decided to use face-cropping. This meant taking a crop over the face in every frame, then restacking the frames back into a video file. It also meant that the frames all had to be the same size and that we didn't stretch the source in-between frames to match this frame size so that no videos had any distortion.
\par
To accomplish this we tried Haar Cascades, then BlazeFace, before settling on RetinaFace \cite{RetinaFace}. The reason why Haar Cascades were unnaceptable for our purposes was that they have a lot of false positives. The reason we didn't find BlazeFace acceptable was that it drew it's bounding boxes pretty inconsistently between frames, causing a lot of jittering. RetinaFace has a slower forward pass than BlazeFace, but still acceptable.
\begin{center}
\begin{figure}[!h]
 \includegraphics[width=1\columnwidth]{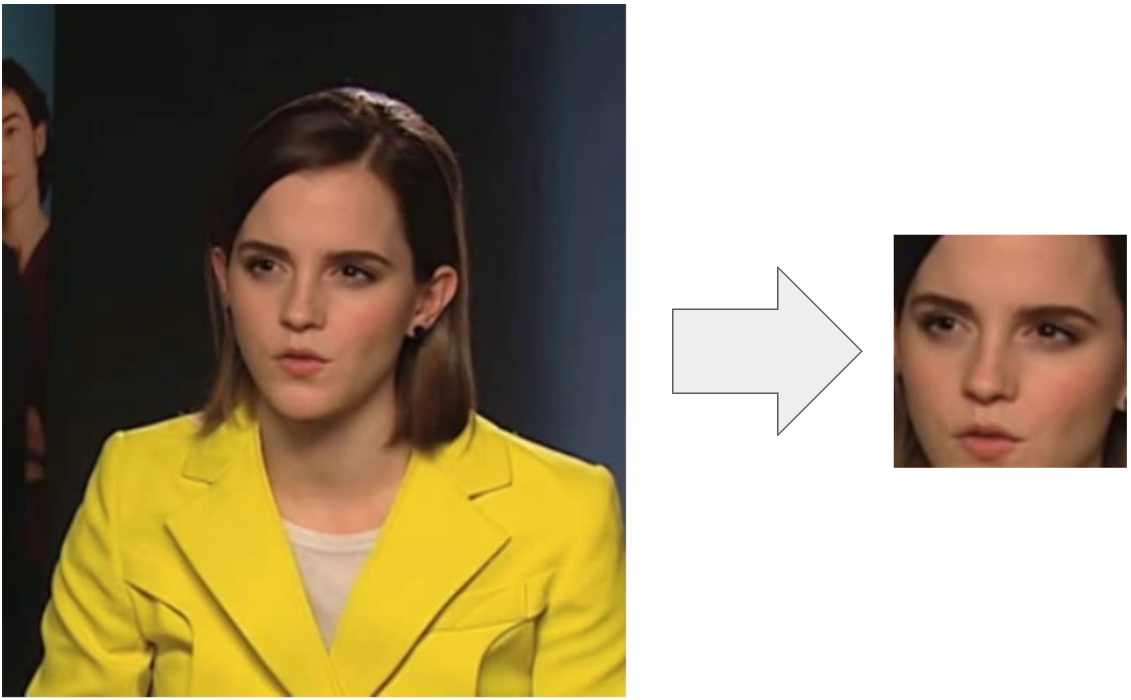}
\caption{Still-frame from Celeb-Df.}
\end{figure}
\end{center}

\subsection{DFT}
One non-temporal classification method was included for a basis of comparison. This method, Unmasking Deepfakes with simple Features \cite{durall2019unmasking}, relies on detecting statistical artifacts in images created by GANs. The discrete Fourier transform of the image is taken, and the 2D amplitude spectrum is compressed into a $300 \times 1$ feature vector with azimuthal averaging. These feature vectors can then be classified with a simple binary classifier, such as the Logistic Regression. This technique can also be used with k-means clustering to effectively classify unlabeled datasets.

\subsection{RCN}
Deepfake videos lack temporal coherence as frame by frame video manipulation produce low level artifacts which manifest themselves as inconsistent temporal artifacts. Such artifacts can be found using a RCN \cite{nguyen2019deep}. The network architecture pipeline consist of a CNN for feature extraction and an LSTM for temporal sequence analysis. A fully connected layer uses the temporal sequence descriptors outputted by the LSTM to classify fake and real videos. Our model consists of two parts: an encoder and a decoder. The encoder comprises of a pre-trained VGG-11 network with batch normalization for extracting features while the decoder is composed of an LSTM with 3 hidden layers followed by 2 fully connected layers. Each video from the dataset was converted into 10 random sequential frames and fed to the network. For each 3D tensor corresponding to the video, features of each frame was found iterating over the time domain and stacked into a tensor which was then fed as input to the decoder.

\subsection{R3D}
3D CNNs \cite{tran2018closer} are able to capture both spatial and temporal information by extracting motion features encoded in adjacent frames in a video. Models like 3D CNN can be relatively shallow compared to 2D image-based CNNs. The R3D network implemented in this paper consists of a sequence of residual networks which introduce shortcut connections bypassing signals between layers. The only difference with respect to traditional residual networks is that the network is performing 3D convolutions and 3D pooling. The tensor computed by the i-th convolutional block is 4 dimensional and has size $N_i \times L \times H_i \times W_i$. $N_i$ is the number of filters used in the $i-th$ block. Just like C3D \cite{tran2018closer}, the kernels have a size of 3 X 3 X 3.  and the temporal stride of conv1 is 1. The input size is 3 x 16 x 112 x 112 where 3 corresponds to the RGB channels and 16 corresponds to the number of consecutive frames buffered. Our implementation follows the 18 layer version of the original paper \cite{hara2017learning} and has a total of 33.17 million weights pretrained on the Kinetics dataset \cite{Kinetics94:online}.

\subsection{ResNet Mixed 3D-2D Convolution}
MC3 builds on the R3D implementation \cite{hara2017learning}.
To address the argument that temporal modelling may not be required over all the layers in the network, the Mixed Convolution architecture \cite{tran2018closer} starts with 3D convolutions and switches to 2D convolutions in the top layers. There are multiple variants of this architecture which involve replacing different groups of 3D convolutions in R3D with 2D convolutions. The specific model used in this paper is MC3 (meaning that layer 3 and deeper are all 2D). Our implementation has 18 layers and 11.49 million weights pretrained on the Kinetics dataset\cite{Kinetics94:online}. The network takes clips of 16 consecutive RGB frames with a size of $112 \times 112$ as input. A stride of $1 \times 2 \times 2$ is used in \texttt{conv1} to downsample, and a stride of $2 \times 2 \times 2$ is used to downsample at \texttt{conv3\_1}, \texttt{conv4\_1}, and \texttt{conv5\_1}.

\subsection{ResNet (2+1)D}
A different approach involves approximating 3D convolution using a 2D convolution followed by a 1D convolution separately. In the R(2+1)D network \cite{tran2018closer} the 3D convolutional filters of size $N_{i-1} \times t \times d \times d$ are replaced with 2D filters of size $N_{i-1} \times 1 \times d \times d$ and $N_i$ temporal convolutional filters of size $M_i \times t \times 1 \times 1$. $M_i$ is a hyperparameter which relates to the dimension of the subspace where the signal is projected between the spatial and temporal convolutions. By separating the 2D and 1D convolutions, more non-linearities are introduced in the network thereby increasing the complexity of functions that can be represented. In addition to this, the factorising convolutions makes optimisation easier resulting in a lower training error. The striding and structure of the network is similar to MC3. The model has 18 layers and 31.30 million parameters pretrained on the Kinetics dataset \cite{Kinetics94:online}.

\subsection{I3D}
One of the highest performing network architectures for spatiotemporal learning is I3D \cite{carreira2017quo}. I3D is an Inflated 3D ConvNet based on 2D ConvNet inflation. The network simply inflates filters and pooling kernels of deep classification ConvNets to 3D, thus allowing spatiotemporal features to be learnt using existing successful 2D architectures pretrained on ImageNet.  During implementation the RGB data is passed to the single-stream I3D network. The architecture used is Inception-V1 \cite{ioffe2015batch} as the base network. Every convolutional layer is followed by a batch
normalization \cite{ioffe2015batch} and a ReLU activation function except for the last convolutional layer. The cropped faces images are resized to 256 x 256 pixels, and then randomly cropped to 224 x 224. The network has 12.29 million weights pretrained on the Charades dataset \cite{Perceptu83:online}. 

\begin{figure}[p]
\begin{tabular}{cc}
\multicolumn{2}{c}{\includegraphics[width=0.9\columnwidth]{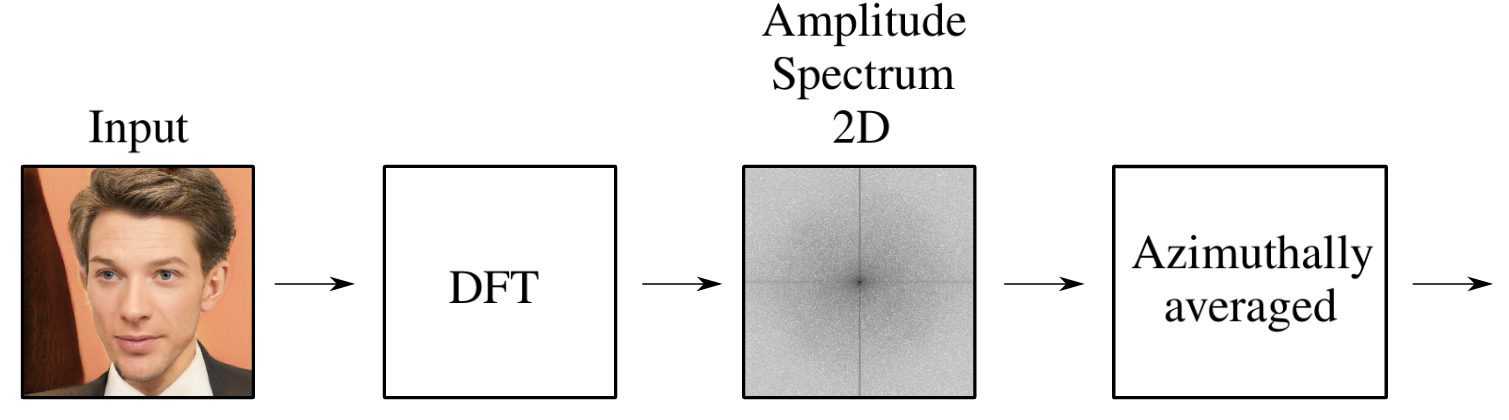}}\\
\multicolumn{2}{c}{\includegraphics[width=0.7\columnwidth]{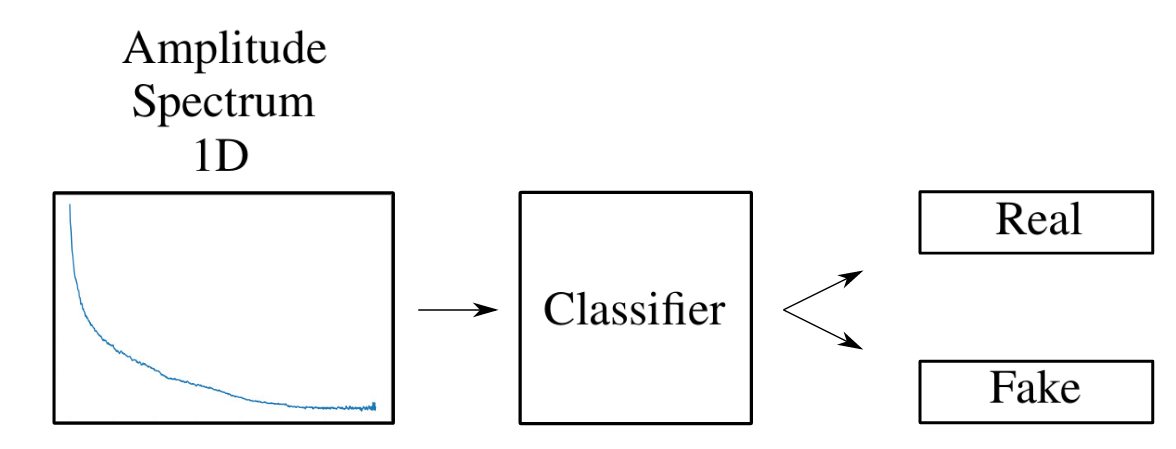}}\\
\multicolumn{2}{c}{(a) DFT data pipeline \cite{durall2019unmasking}}\\[6pt]
  \includegraphics[width=0.25\columnwidth, height=4cm]{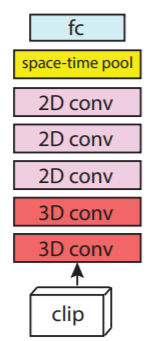} &   \includegraphics[width=0.25\columnwidth, height=4cm]{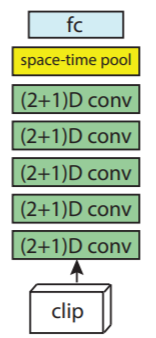} \\
(b) MCx \cite{tran2018closer}  & (c) R(2+1)D \cite{tran2018closer} \\[6pt]
 \includegraphics[width=0.25\columnwidth, height=4cm]{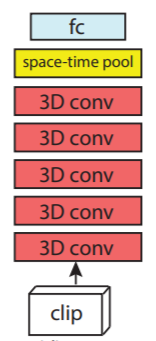} &  \includegraphics[width=0.45\columnwidth]{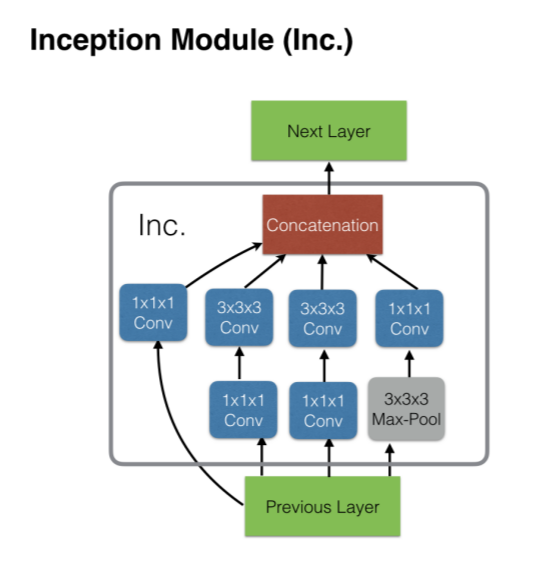}  \\
(d) R3D \cite{tran2018closer} & (e) Inception Block (I3D) \cite{carreira2017quo} \\[6pt]
\multicolumn{2}{c}{\includegraphics[width=0.9\columnwidth]{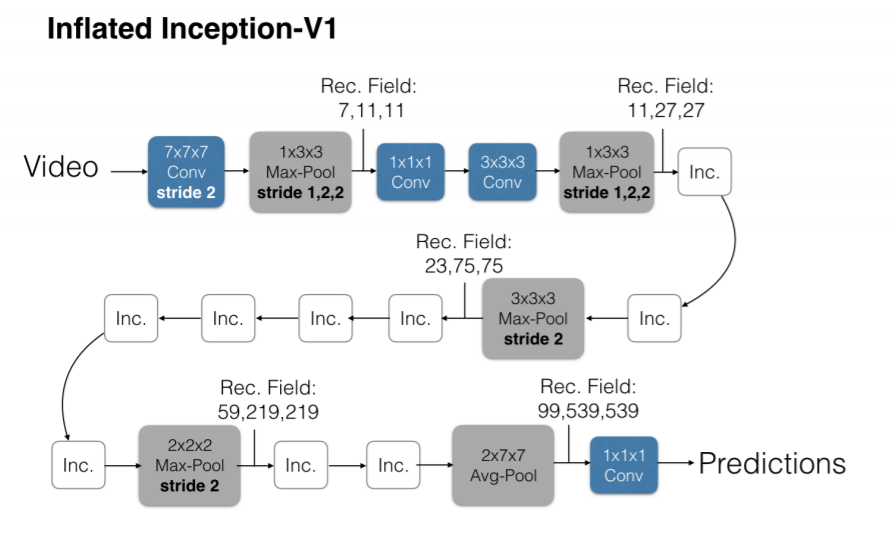}}\\
\multicolumn{2}{c}{(f) Inflated Inception V1 (I3D) \cite{carreira2017quo}}\\

\multicolumn{2}{c}{\includegraphics[width=0.8\columnwidth]{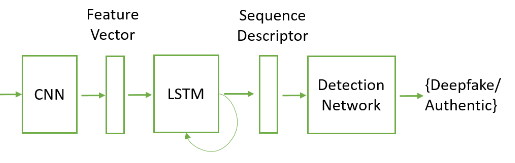}
}\\
\multicolumn{2}{c}{(g) RCNN \cite{nguyen2019use}}
\end{tabular}
\caption{Network architectures investigated in this paper.}
\label{fig:architectures}
\end{figure}